\documentclass[runningheads]{llncs}
\usepackage{graphicx}

\usepackage{tikz}
\usepackage{comment}
\usepackage{amsmath,amssymb} 
\usepackage{color}

\usepackage{epsfig}
\usepackage{graphicx}
\usepackage{amsmath}
\usepackage{amssymb}
\usepackage{bbm}
\usepackage{epsfig}
\usepackage{graphicx}
\usepackage{amsmath}
\usepackage{amssymb}
\usepackage{grffile}
\usepackage{algorithm,algcompatible,amsmath}
\usepackage{subfigure}
\usepackage{multirow}
\usepackage{booktabs}
\usepackage{enumitem}
\usepackage{mdwlist} 
\usepackage{fixltx2e}
\usepackage{lipsum}
\usepackage{bm}

\usepackage{color,xcolor}
\usepackage{dcolumn}
\usepackage{hyperref}

\usepackage[accsupp]{axessibility}  

\usepackage{float}
\usepackage{graphicx}

\begin{document}
\pagestyle{headings}
\mainmatter
\def\ECCVSubNumber{3147}  

\title{A Large-scale Multiple-objective Method for Black-box Attack against Object Detection} 



\titlerunning{GARSDC}
%
\author{Siyuan Liang\inst{1,2} \and
Longkang Li\inst{3} \and
Yanbo Fan\inst{4} \and
Xiaojun Jia\inst{1,2} \and
Jingzhi Li\inst{1,2, \thanks{Coressponding Author}} \and
Baoyuan Wu\inst{3,\star} \and
Xiaochun Cao \inst{5}}
%
%

\authorrunning{Siyuan Liang, et al.}
\institute{State Key Laboratory of Information Security,
Institute of Information Engineering, Chinese Academy of Sciences, Beijing, China  \and  
School of Cyber Security, University
of Chinese Academy of Sciences, Beijing, China \and
School of Data Science, Secure Computing Lab of Big Data, The Chinese University of Hong Kong, Shenzhen, China  \and
Tencent AI Lab, Shenzhen, China \and
School of Cyber Science and Technology, Shenzhen Campus, Sun Yat-sen University, Shenzhen, China \\
\email{\{liangsiyuan, jiaxiaojun, lijingzhi\}@iie.ac.cn};
\email{\{lilongkang, wubaoyuan\}@cuhk.edu.cn};
\email{fanyanbo0124@gmail.com};
\email{caoxiaochun@mail.sysu.edu.cn} 
}

\maketitle
\begin{abstract}

Recent studies have shown that detectors based on deep models are vulnerable to adversarial examples, even in the black-box scenario where the attacker cannot access the model information. Most existing attack methods aim to minimize the true positive rate, which often shows poor attack performance, as another sub-optimal bounding box may be detected around the attacked bounding box to be the new true positive one. To settle this challenge, we propose to minimize the true positive rate and maximize the false positive rate, which can encourage more false positive objects to block the generation of new true positive bounding boxes. It is modeled as a multi-objective optimization (MOP) problem, of which the generic algorithm can search the Pareto-optimal. However, our task has more than two million decision variables, leading to low searching efficiency. Thus, we extend the standard \textbf{G}enetic \textbf{A}lgorithm with \textbf{R}andom \textbf{S}ubset selection and \textbf{D}ivide-and-\textbf{C}onquer, called GARSDC, which significantly improves the efficiency. Moreover, to alleviate the sensitivity to population quality in generic algorithms, we generate a gradient-prior initial population, utilizing the transferability between different detectors with similar backbones. Compared with the state-of-art attack methods, GARSDC decreases by an average 12.0 in the mAP and queries by about 1000 times in extensive experiments. Our codes can be found at https://github.com/LiangSiyuan21/\\GARSDC.

\keywords{Adversarial Learning, Object Detection, Black-box Attack}
\end{abstract}

\section{Introduction}

 
 With the development of deep learning, object detection~\cite{zhao2019object,tan2020efficientdet,kong2020foveabox,joseph2021towards} has been widely applied in many practical scenarios, such as autonomous driving~\cite{levinson2011towards}, face recognition~\cite{jafri2009survey}, industrial detection~\cite{hu2018survey}, etc. In object detection, the true positive object refers to the positive object correctly and the false positive object refers to the negative object that is incorrectly marked as positive object. 
\begin{figure}[t]
    \setlength{\abovecaptionskip}{-0.5cm}
	\begin{center}
		\includegraphics[width=0.85\linewidth]{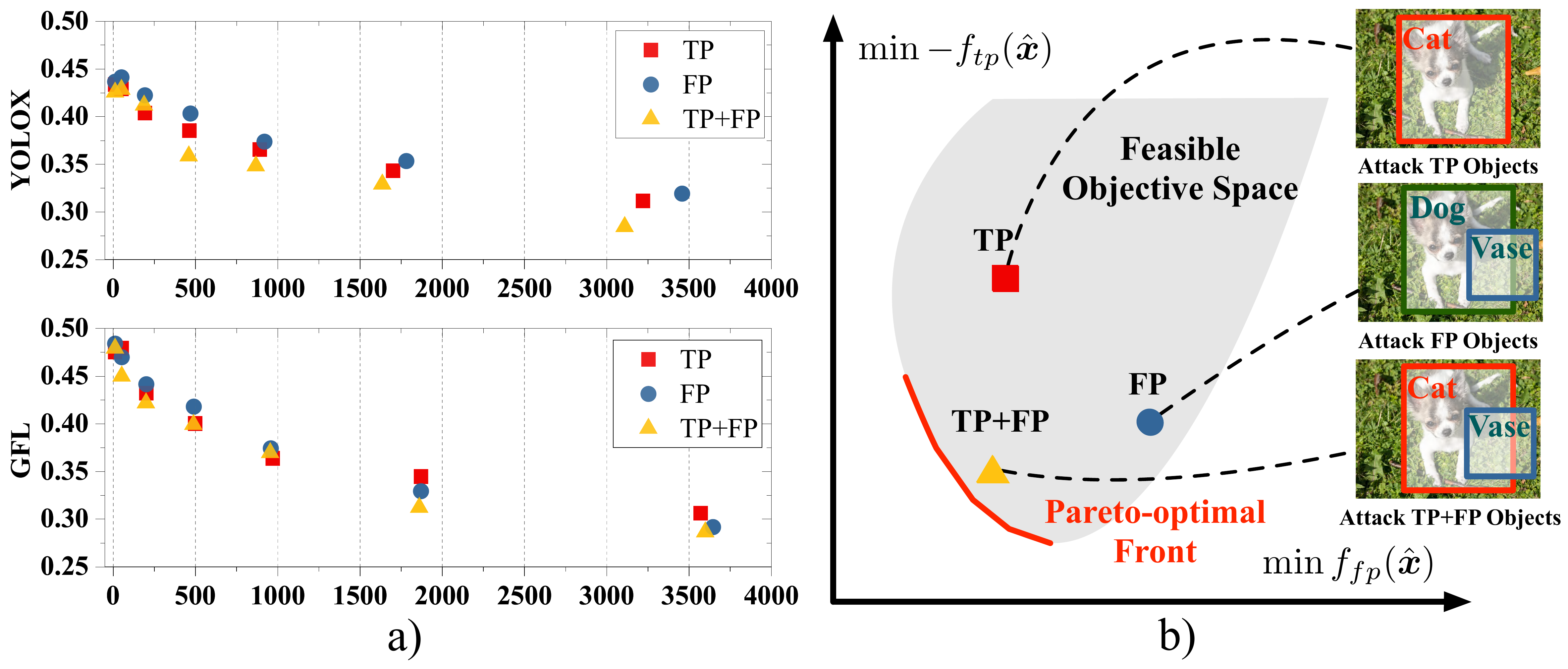}
	\end{center}
	\caption{$\bm{a)}$ We show the results of attacking two different models using three objective functions(TP, FP, `TP+FP'). Experiments show that using `TP+FP' can decrease the most mAP and reduce most queries. $\bm{b)}$ We show the difference of three objective optimizations, and the total consideration of `TP+FP' makes the solution closer to the Pareto-optimization front.}
	\label{fig1}
\end{figure}

Recently, the adversarial attack and defense around deep learning have received extensive attention~\cite{jia2022adversarial,bai2020targeted,wang2021Enhancing,wu2020wider,9807638}. Existing attacks against object detection misclassify the true positive objects from the model, which leads to attack failure. The reason is that another sub-optimal bounding box can replace the attacked bounding box successfully as the new true positive object. Another train of thought, increasing the false positive objects is also an effective attack method in some scenarios. For example, many false objects can obscure the significance of the positive object and lead to autonomous driving system~\cite{shim2015autonomous} crashes. Therefore, we believe that is critical and desirable to simultaneously optimize true positive and false positive objects recognized by the detector for the following reasons. Firstly, optimizing the objective function with two aspects can expand the attack scenarios. Secondly, we minimize the true positive rate and maximize the false positive rate, which increases false positive objects to block the generation of the true positive object. Thirdly, considering the attack target comprehensively helps decrease the mAP. As shown in Fig.~\ref{fig1}, through experiments on YOLOX and GFL models, we prove that optimizing the true positive or false positive objects can attack the detector successfully, and optimizing both of them will achieve better attack performance. Attacks on existing detectors are not comprehensive due to a lack of consideration of false positive objects.

Inspired by the statements above, we reformulate the adversarial attack~\cite{wang2021universal,wang2021dual,liu2020bias} against object detection as a large-scale multi-objective optimization problem (MOP)~\cite{deb2014multi} to decrease true positive objects and increase false positive objects. Our interest focuses on the black-box settings. A large-scale MOP under the black-box setting mainly faces three challenges. Firstly, the conflict between multiple objectives makes it almost impossible to find a solution that optimizes all objectives simultaneously. Secondly, decision variables are extremely large-scaled due to the consistent dimensions of adversarial samples and images (more than two million). Nevertheless, the existing optimization algorithms w.r.t. MOP have poor scalability, and optimizing decision variables with more than one million is especially tough~\cite{hong2021evolutionary}. Thirdly, black-box attacks should reduce queries while increasing the attack rate. To address the above challenges, we use genetic algorithms to find optimal trade-off solutions for MOP, called Pareto-optimal solutions~\cite{deb2005searching}. A genetic algorithm can approximate the entire set of Pareto-optimal in a single run and does not make specific assumptions about the objective functions, such as continuity or differentiability~\cite{hong2021evolutionary}. To settle the poor scalability, we propose a \textbf{G}enetic \textbf{A}lgorithm based on \textbf{R}andom \textbf{S}ubset selection and a \textbf{D}ivide-and-\textbf{C}onquer 
algorithm to optimize large-scale decision variables, named as GARSDC. This method aims to transform the original search space of MOP using dimensionality reduction and divide-and-conquer, improving the optimization algorithm's searchability by rebalancing exploration and exploitation. The genetic algorithm is sensitive to the population. Thus, we use gradient-based perturbations as the initial population. Moreover, we analyze more than 40 object detection backbones and find out that the perturbation is transferring well in the same backbone. Thus, we generate the chain-based and skip-based perturbations as a mixed initial population with transferability. By combining transfer and query-based attacks, our method substantially decreases the mAP and queries on eight representative detectors than the state-of-the-art method. This paper has the following contributions to the three-fold:
\begin{enumerate}
 \item We model the adversarial attack problem against object detection as a large-scale multi-objective optimization, which can expand the attack scenarios and help understand the attack mechanism against object detection. Experiments show that this comprehensive modeling helps to decrease the mAP.
 \item We design a genetic algorithm based on random subset selection and divide-and-conquer methodology for solving Pareto-optimal solutions, called GARSDC, which improves the searchability of GA by rebalancing the exploration and exploitation of the optimization problem. We generate chain-based and skip-based perturbations as a mixed initial population with gradient-prior, increasing population diversity and improving the algorithm's efficiency.
 \item A large number of attack experiments based on different backbone detectors demonstrate the effectiveness and efficiency of GARSDC. Compared with the state-of-art PRFA algorithm, GARSDC reduces by an average 12.0 in the mAP and queries by about 1000 times.	
\end{enumerate}

\section{Related Work}

\subsection{Object Detection Based on Deep Learning}

In recent years, the latest progress of object detectors mainly focuses on three aspects: Firstly, the improvement of the backbone network, detectors based on different backbones have produced significant differences in accuracy and inference speed. Standard models include SSD~\cite{liu2016ssd} based on VGG16, Centernet~\cite{duan2019centernet} based on ResNet18 and YOLOX~\cite{ge2021yolox} based on yolo-s network. Most models are based on ResNet~\cite{ren2015faster} and FPN~\cite{lin2017feature} series architecture, e.g., Cascade R-CNN~\cite{cai2018cascade}, Atss~\cite{zhang2020bridging}, Fcos~\cite{tian2019fcos}, and Freeanchor~\cite{zhang2019freeanchor}. Secondly, combining the learning of instance segmentation, such as segmentation annotation in Mask R-CNN~\cite{he2017mask} and switchable atrous convolution in Detectors~\cite{qiao2021detectors}. Thirdly, improvements of localization, such as GFL~\cite{li2020generalized} based on generalized focal loss. Our research finds that detectors that focus on different improvements have significant differences in transfer attacks. In addition, compared with detectors based on different backbones, detectors with the same backbone structure are less challenging to transfer, and this phenomenon brings excellent inspiration to our model selection for transfer attacks.

\subsection{Black-box Adversarial Attack}
Generally speaking, black-box adversarial attacks can be divided into transfer attacks, decision-based attacks, and score-based attacks. The transfer attack, also known as the local surrogate model attack, assumes that the attacker has access to part of the training dataset to train the surrogate model, including adaptive black-box attack~\cite{papernot2017practical} and data-free surrogate model attack~\cite{zhou2020dast}. The score-based attack allows the attacker to query the classifier and get probabilistic of the model prediction. Representative methods include the square attack~\cite{andriushchenko2020square} based on random search and the black box attack based on transfer prior. Decision-based attacks~\cite{chen2020hopskipjumpattack} can obtain less information than the above, allowing the attacker to accept label outputs instead of probabilities. By analyzing the architectural characteristics of the object detector, we improve the efficiency and accuracy of score-based attacks according to the gradient-prior.

\subsection{Adversarial Attack against Object Detection}

The existing adversarial attack methods for object detection are mainly based on white-box attacks, and the attacker implements the adversarial attack by changing the predicted label of the true positive object. DAG~\cite{xie2017adversarial}, and CAP~\cite{zhang2020contextual} mainly implement adversarial attacks by fooling the RPN network of two-stage detectors in terms of the types of attack detectors. To increase the generality of the attack algorithm, UEA~\cite{wei2018transferable} and TOG~\cite{chow2020adversarial} exploit transferable adversarial perturbations to attack both the one-stage detector and the two-stage detector simultaneously. PRFA~\cite{liang2022parallel} first proposes a query-based black-box attack algorithm to fool existing detectors using a parallel rectangle flipping strategy. This method also provides a baseline for target detection query attacks. Our proposed algorithm not only surpasses the state-of-the-art algorithm PRFA but also attacks more representative detection models, comprehensively evaluating the robustness of existing detectors.

\section{Method}

\subsection{Simulating Adversarial Examples Generating by MOP}
We firstly introduce the background of MOP, including problem definition, non-dominant relations, and Pareto solutions. A MOP problem can be mathematically modeled as:
\begin{equation}
\begin{aligned}
	\min F(\hat{\boldsymbol{x}})=(f_1(\hat{\boldsymbol{x}}),...,f_K(\hat{\boldsymbol{x}})),\hat{\boldsymbol{x}}=(\hat{x}_1,...\hat{x}_D)\in \rm{\Omega},
\end{aligned}
\label{eq1}	
\end{equation}
where there are $D$ decision variables with respect to the decision vector $\hat{\boldsymbol{x}}$, the objective function $F:\rm{\rm{\Omega}}\rightarrow \mathbf{R}^{K}$ includes $K$ objective functions, $\rm{\Omega}$ and $K$ represent the decision and objective spaces.  Generally speaking, when the $K \geq 2$ and $D \geq 100$, when call this MOP as a large-scale MOP.

\noindent\textbf{Definition 1.} Given two feasible solutions $\hat{\boldsymbol{x}}_1$, $\hat{\boldsymbol{x}}_2$ and their objective functions $F(\hat{\boldsymbol{x}}_1)$, $F(\hat{\boldsymbol{x}}_2)$, $\hat{\boldsymbol{x}}_1$ dominates $\hat{\boldsymbol{x}}_2$ (denoted $\hat{\boldsymbol{x}}_1 \prec \hat{\boldsymbol{x}}_2$) if and only if $\forall i \in \{1,...,K\}, f_i(\hat{\boldsymbol{x}}_1) \leq f_i(\hat{\boldsymbol{x}}_2)$ and $\exists j \in \{1,...,K\}, f_j(\hat{\boldsymbol{x}}_1)< f_j(\hat{\boldsymbol{x}}_2)$.

Definition 1 describes the dominance relation in the MOP.

\noindent\textbf{Definition 2.} A solution $\hat{\boldsymbol{x}}^{*}$ is Pareto-optimal solution if and only if there exists no $\hat{\boldsymbol{x}}_1 \in \rm{\Omega}$ such that $F(\hat{\boldsymbol{x}}_1) \prec F(\hat{\boldsymbol{x}}^{*})$. We name the set of all Pareto-optimal solutions as the Pareto Set and the corresponding objective vector set as the Pareto front. 

Given a large-scale MOP, we describes the Pareto solution in Definition 2. We have a clean image $\boldsymbol{x}$ containing a set of $M$ recognition objects $\mathcal{O}$ , that is, $\mathcal{O}=\{o_1,...,o_M\}$. Each recognition objects $o_i$ is assigned a groud-truth class $o_{i}^c \in \{1,...,C\}$, $i \in \{1,...,M\}$. $C$ is the number of class, the $C=81$ in the MS-COCO. The object detector $H$ predict $N$ objects as the predicted objects $\mathcal{P}=H(\boldsymbol{x})$ and the corresponding classes $\mathcal{P}^{c}$ in the clean image $\boldsymbol{x}$. However, limited by training datasets and complex scenes, the objects $\mathcal{P}$ predicted by the detector are not always consistent with the recognized objects $\mathcal{O}$. We define the true positive object as follows: there is a only one object $o_i$ such that the intersection of union between $p_j$ and $o_i$ greater than 0.5 and $p_j^{c}$ is same with $o_i^{c}$, otherwise it is a false positive object~\cite{everingham2008pascal}. Thus, we decompose the predicted $\mathcal{P}$ objects as true positive objects $\mathcal{TP}$ and false positive objects $\mathcal{FP}$ by recognition objects $\mathcal{O}$. The $|\mathcal{P}|=|\mathcal{TP}|+|\mathcal{FP}|$. Previous adversarial examples $\hat{\boldsymbol{x}}$ with a small $\boldsymbol{\delta}$ attack the detector by reducing the true positive objects. Fig.~\ref{fig1} a) show that attacking the false positive objects alone can also attack the detector. Therefore, we model the adversarial attack as a large-scale MOP: reducing the true positive objects and increasing false positive object. The objective function can be represented as:
\begin{equation}
\begin{aligned}
	F(\hat{\boldsymbol{x}}, H(\boldsymbol{x}))=\min(-f_{tp}(\hat{\boldsymbol{x}},\mathcal{P}),f_{fp}(\hat{\boldsymbol{x}},\mathcal{P})) , s.t. \  \hat{\boldsymbol{x}}=(\boldsymbol{x}+\boldsymbol{\delta}) {\in} \rm{\Omega}, \ ||\hat{\boldsymbol{x}}-\boldsymbol{x}||_n {\leq} \epsilon,
\end{aligned}
\label{eq2}
\end{equation}
where $n$ denotes norm. We solve the problem in the weighting method~\cite{zadeh1963optimality}. The Eq.~\eqref{eq2} can be written as:
\begin{equation}
\begin{aligned}
	F(\hat{\boldsymbol{x}}, H(\boldsymbol{x})) = \min(w_1*(-f_{tp}(\hat{\boldsymbol{x}},\mathcal{P})) &+ w_2* f_{fp}(\hat{\boldsymbol{x}},\mathcal{P}))
	\\ s.t. \ \hat{\boldsymbol{x}}=(\boldsymbol{x}+\boldsymbol{\delta}) {\in} \rm{\Omega}, \  ||\hat{\boldsymbol{x}}-\boldsymbol{x}||_n& {\leq} \epsilon, 
\end{aligned}
\label{eq3}
\end{equation}
where $w_i \geq 0$.We use the CW loss~\cite{carlini2017adversarial} as attack functions $f_{tp}$ or $f_{fp}$:
\begin{equation}
f_{\{tp,fp\}}(\hat{\boldsymbol{x}},\mathcal{P}) = \sum_{i \in \{\mathcal{TP},\mathcal{FP}\}}^{|\mathcal{P}|}\left(\max_{l\neq c}(f_{\{tp,fp\}}(\hat{\boldsymbol{x}},p_i)_{l})-f_{\{tp,fp\}}(\hat{\boldsymbol{x}},p_i)_c\right),
\label{eq4}
\end{equation}
where $i \in  \mathcal{TP}$ represents the $i$-th predicted box, $p_i$ is the true positive object in the predicted boxes $\mathcal{P}$. In Eq.~\eqref{eq3}, we aim to make the labels of true positive objects wrong and protect the false positive objects. Thus, we treat $f_{tp}$ and $f_{fp}$ as untargeted and targeted attacks, respectively.

\begin{figure}[t]
    \setlength{\abovecaptionskip}{-0.3cm}
	\begin{center}
		\includegraphics[width=0.85\linewidth]{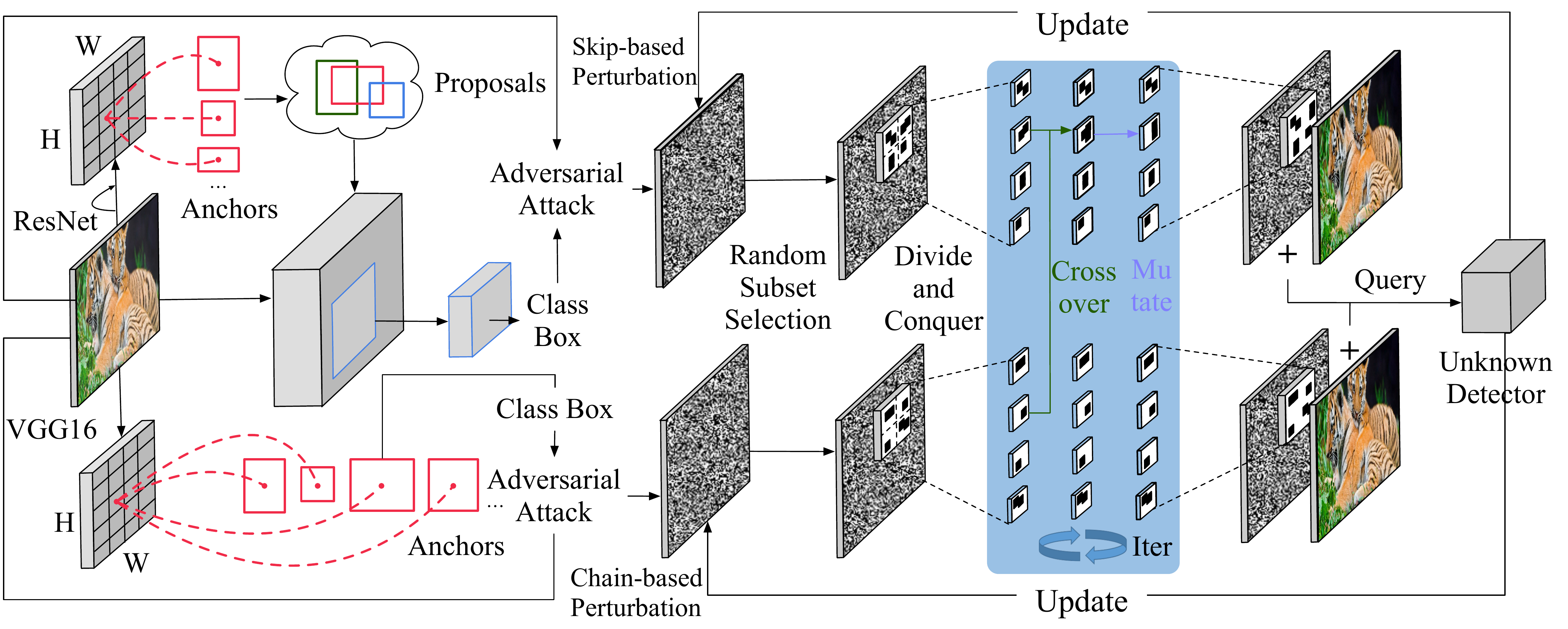}
	\end{center}
	\caption{To optimize multi-objective problems, we propose a \textbf{G}enetic \textbf{A}lgorithm based on \textbf{R}andom \textbf{S}ubset selection and a \textbf{D}ivide-and-\textbf{C}onquer algorithm (GARSDC). The basic flow of the GARSDC attack is shown above, which combines the transfer-based and the query-based attacks against the black-box model.}
	\label{fig2}
\end{figure}

\subsection{Generating Adversarial Examples by Genetic Algorithm}
Since the genetic algorithm is based on the nature of the population and does not require additional assumptions (continuous or differentiable) for objective functions, the genetic algorithm can gradually approximate the Pareto-optimal solution in the single queries~\cite{hong2021evolutionary}. We choose the genetic algorithm, which only uses the fitness function to evaluate individuals in the population and search the best individual as the adversarial perturbation. We define the initial population $\Delta^0$ containing $P$ individuals as $\Delta^0=\{\boldsymbol{\delta}_{1}^0,...,\boldsymbol{\delta}_{p}^0\}$ and the $p$-th individual fitness $P(\boldsymbol{x}+\boldsymbol{\delta}_p)=F(\boldsymbol{x}+\boldsymbol{\delta}_p)$. The population is iterating in the direction of greater individual fitness. Generating the $i$-th population $\Delta^{i}$ mainly relies on crossover and mutation. The greater the individual fitness, the more likely it is to be saved as the next population. For example, if $P(\boldsymbol{\delta}_1^{i})>P(\boldsymbol{\delta}_2^{i})$, then the next individual $\boldsymbol{\delta}_2^{i+1}$ will inherit some features (crossover) of $\boldsymbol{\delta}_1^{i}$ and mutate. In Fig.~\ref{fig2}, the transfer attack generate the initial population $\Delta^0$. The iteration stopping condition of population iteration is when reaching the maximum iteration, or the fitness is greater than a certain value. The optimal solution of the population is the individual with the greatest fitness, that is, the adversarial perturbation we need. Since our decision space is too large ($weight{*}height{*}channel$ exceeds millions), it is difficult for general genetic algorithms to converge in limited queries. Next, we will introduce the improved genetic algorithm from the gradient-prior initial population, random subset selection, and divide-and-conquer algorithm.

\begin{figure}[t]
    \setlength{\abovecaptionskip}{-0.1cm}
	\begin{center}
		\includegraphics[width=0.8\linewidth]{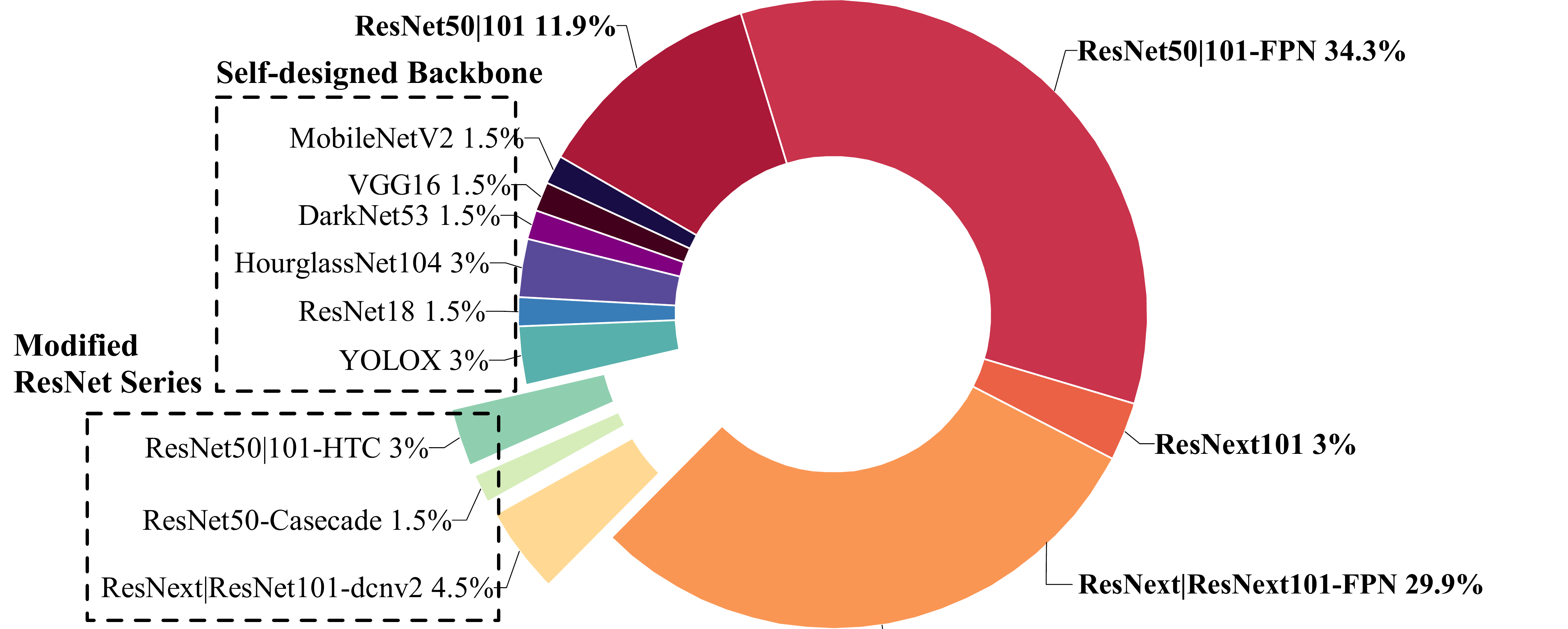}
	\end{center}
	\caption{The investigated results of detectors based on different backbone networks.}
	\label{fig3}
\end{figure}

\subsection{Mixed Initial Population Based on Gradient-prior}

An excellent initial population can help the genetic algorithm converge more quickly, so finding a suitable initial population for the black-box detector is critical. Although the QAIR~\cite{li2021qair} algorithm estimates the gradient of the adversarial perturbation by stealing the image retrieval system, the cost of model stealing for the detector is too high. Because the detectors are diverse and the dataset for object detection relies on enormous annotations. Intuitively, we can generate adversarial perturbations with well transferability as an initial population against the detector.

Inspired by that transferable perturbation in image classification can attack different feature networks, we analyze more than 40 deep model-based object detectors and classify their backbone network types. In Fig.~\ref{fig3}, detectors based on backbone networks belong to ResNet, ResNet-FPN~\cite{lin2017feature}, and their derivatives, e.g., ResNeXt~\cite{xie2017aggregated}, account for more than 80\%. We call these backbone networks the ResNet series. The other two types of backbone networks are based on the modified ResNet series, e.g., Detectors~\cite{qiao2021detectors}, or self-designed networks, such as YOLOX~\cite{ge2021yolox} based on yolo-s. Therefore, we can roughly divide the current network architecture into three categories and attack against the ResNet series, the modified ResNet series, and self-designed networks. Although detection models vary widely in network structures, both of them use the cross-entropy loss of the prediction boxes. We can implement an adversarial attack by maximizing the cross-entropy loss of all prediction boxes, and the objective function is as follows:
\begin{equation}
\begin{aligned}
	D(\boldsymbol{x}+\boldsymbol{\delta})=\sum_{i=1}^{|\mathcal{P}|}\sum_{j=1}^{|C|}y_{ij}*\log(\boldsymbol{c}_{ij})
\end{aligned}
\label{eq5}
\end{equation}
where $y_{ij}$ is one when detector $H$ classify the $i$-th prediction box into the $j$-th category, otherwise $y_{ij}$ is zero. $\boldsymbol{c}$ represents the classification probability of the $i$-th prediction box. We can attack the Eq.~\eqref{eq5} using an off-the-shelf transfer attack algorithm, such as TI-FGSM~\cite{dong2019evading}.


Although the input can be randomly initialized by adding noise to the clean image, adversarial perturbations based on the same detector lack diversity. To accelerate the genetic algorithm convergence, the diversity of individuals is essential. Therefore, we attack detectors with different backbone networks to generate individuals with differences, and we call this population composed of sexual individuals as the mixed initial population based on gradient-prior. Specifically, we respectively select the initial individuals attacked by the VGG16-based and ResNet-based detectors. In essence, VGG16 and ResNet are different because ResNet is a backbone network with a skip-connection structure, and VGG16 is a chain structure. We refer to the different individuals generated by these two networks as skip-based perturbation and chain-based perturbation.

\subsection{Random Subset Selection}

The variable space($weight{*}height{*}channel$) of the perturbation exceeds millions, and the intuitive idea for solving the large-scale MOP is to decompose high-dimensional decision variables into many low-dimensional sub-components and assign MOP to sub-components through specific strategies, which solves the MOP indirectly by optimizing a portion of the MOP. We will introduce the random subset selection for sub-components and the corresponding MOP decomposition strategy. 

Since there are many decision variable combinations for sub-component selection, it is unrealistic to traverse all combinations in a limited number of queries. We use a random subset selection algorithm to sample sub-component in the decision space. We use the index vector $\boldsymbol{s}\in \{0, 1\}^{D}$ for random subset selection. If $s_i=1$ the $i$-th element of $\delta$ is selected and $s_i=0$ otherwise. Square attack~\cite{andriushchenko2020square} achieves good black-box attack performance by generating square-shaped adversarial patches through random search in the image. It is feasible to select adversarial patches randomly to attack the detector. This process can be regarded as sampling the sub-component $\boldsymbol{\delta}[\boldsymbol{s}]$ in the decision space $\boldsymbol{\delta}$. 

In section 3.2, we introduce the individual fitness. However, the computation of individual fitness is for the overall adversarial perturbation $\boldsymbol{\delta}$ rather than the sub-component $\boldsymbol{\delta}[\boldsymbol{s}]$. We can easily computer the coordinates $s^{B}$ of sub-component $\boldsymbol{\delta}[\boldsymbol{s}]$ by using $\boldsymbol{s}$, then we assign the predicted boxes $\mathcal{P}$ to the sub-component and calculate the fintess:
\begin{equation}
f_{\{tp,fp\}}(\hat{\boldsymbol{x}},\mathcal{P},s^B) = \sum_{i \in \{\mathcal{TP},\mathcal{FP}\}}^{|\mathcal{P}|}(\max_{l\neq c}(f_{\{tp,fp\}}(\hat{\boldsymbol{x}},p_i)_{l})-f_{\{tp,fp\}}(\hat{\boldsymbol{x}},p_i)_c)*IoU(p_{i}, s^B),
\label{eq6}
\end{equation}
where $IoU(a,b)$ denotes intersection over Union between $a$ and $b$. The sub-component fitness $S(\boldsymbol{\delta}[\boldsymbol{s}])$ defines as follows:
\begin{equation}
\begin{aligned}
S(\boldsymbol{\delta}[\boldsymbol{s}]) = S(\boldsymbol{x}+\boldsymbol{\delta}, H(\boldsymbol{x}), s^B) = S(\hat{\boldsymbol{x}}, H(\boldsymbol{x}), s^B&)
\\=\min (-f_{tp}(\hat{\boldsymbol{x}}, \mathcal{P}, s^B), f_{fp}(\hat{\boldsymbol{x}}, \mathcal{P}, s^B)).	
\end{aligned}
 \label{eq7}
\end{equation}

We can judge the relationship between the current sub-component and the predicted boxes by calculating the fitness of individual and sub-component. If there is no connection, we discard the current sub-component. Although PRFA has a similar operation that randomly searches for sub-components in the search space, our method has the following innovations: Firstly, we do not need a priori-guided dimension reduction but instead search the image globally, which can circumvent the risk of that the prior is terrible; secondly, we can use the window fitness to help judge whether the sub-components of random search are helpful for optimization.

\subsection{Divide-and-Conquer Algorithm} Although the decision variables are greatly reduced by randomly selecting the sub-component $\boldsymbol{\delta}[\boldsymbol{s}]$, we can still use the divide-and-conquer method for the sub-component $\boldsymbol{\delta}[\boldsymbol{s}]$ to improve the optimization. In Fig.~\ref{fig2}, we show the divide-and-conquer process of sub-component $\boldsymbol{\delta}[\boldsymbol{s}]$. Suppose we decompose the index vector $\boldsymbol{s}$ into $i$ parts, that is $\boldsymbol{s}=\{\boldsymbol{s}_1,...,\boldsymbol{s}_i\}$. We can perform the genetic algorithm in the $i$-th part of the index vector $\boldsymbol{s}_i$ to find a subset $\boldsymbol{u}_{i}$ with a budget $z$. Assuming that we have two individuals $\boldsymbol{\delta_1}$ and $\boldsymbol{\delta_2}$, we can calculate the fitness of sub-components $S(\boldsymbol{\delta_1}[\boldsymbol{s}_{i}])$ and $S(\boldsymbol{\delta_2}[\boldsymbol{s}_{i}])$, respectively. If $S(\boldsymbol{\delta_1}[\boldsymbol{s}_{i}])>S(\boldsymbol{\delta_2}[\boldsymbol{s}_{i}])$, the $\boldsymbol{\delta_2}[\boldsymbol{s}_{i}]$ gets the feature from $\boldsymbol{\delta_1}[\boldsymbol{s}_{i}]$(cross over) and mutates. The $\boldsymbol{\delta_2}[\boldsymbol{s}_{i}]$ updates and gets the new subset $\boldsymbol{u}_i$ from $\boldsymbol{s}_i$. Merge all new subsets into a set $U=\cup_{j=1}^i \boldsymbol{u}_i$. And we find the $\boldsymbol{u}_{i+1}$ in the set $U$. Then, we return the best individual fitness and the corresponding sub-component $\boldsymbol{\delta}[\boldsymbol{u}_{best}]$.

We will analyze the approximation of the divide-and-conquer algorithm in Lemma 1. For $1 \leq j \leq i $, let $\boldsymbol{b}_j \in \arg \max_{\boldsymbol{u} \subseteq \boldsymbol{s}_j:|\boldsymbol{u}|\leq z}P(\boldsymbol{\delta}[\boldsymbol{u}])$ denotes an optimal subset of $\boldsymbol{s}_i$. 

\noindent\textbf{Lemma 1.}~\cite{qian2018distributed} For any partition of $\boldsymbol{s}$, it holds that
\begin{equation}
\label{eq8}
	\max \{P(\boldsymbol{\delta}[\boldsymbol{b_j}])|1 \leq j \leq i \} \geq \{\alpha/i, \gamma_{\emptyset, z}/z\} * OPT,
\end{equation}
where the $\gamma$- and $\alpha$ are submodularity ratios~\cite{qian2017subset}. The $OPT$ denote the value of the objective function in Eq.~\eqref{eq3} For any subset $\boldsymbol{u} \subseteq \boldsymbol{s}_j$, there exists another item, the inclusion of which can improve the individual fitness by at least a proportional to the current distance the best solution~\cite{qian2016parallel}. Then, we can get the approximation performance of divide-and-conquer method for random subset selection with monotone objective functions. For random subset selection with a monotone objective function $P$, our algorithm using $\mathbb{E}[\max\{T_j|1 \leq j \leq i\}]=O(z^2|\boldsymbol{s}|(1+\log i))$ finds a subset $\boldsymbol{u}$ with $\boldsymbol{u} \leq z$ and
\begin{equation}
\label{eq9}
P(\boldsymbol{\delta}[\boldsymbol{u}]) \geq (1-e^{-\gamma_{\min}}) * \max \{\alpha/i, \gamma_{\emptyset, z}/z\} * OPT,
\end{equation}
where $\gamma_{\min}=\min_{\boldsymbol{u} \subseteq \boldsymbol{s}_j:|\boldsymbol{u}|=z-1}\gamma_{\boldsymbol{u}, z}$. 

The above is the complete GARSDC algorithm. Due to the limitation of the paper space, we put the proofs of Eq.~\eqref{eq8} and Eq.~\eqref{eq9} and the algorithm flow of GARSDC in the supplementary materials.

\section{Experiments}
\subsection{Experiment Settings}
\textbf{Dataset and Evaluation.} The current object detectors use the MS-COCO dataset as a benchmark for evaluating performance. For a fair comparison with PRFA, we adopt the same experimental setup as PRFA and use a part of the MS-COCO validation set as the attack image. We use the evaluation matrix (mAP) to evaluate the detection results of the detector on adversarial examples. The lower the mean average precision, the better the attack effect. We evaluate the efficiency of the algorithm using average queries. Under the limit of 4,000 queries per image, the lower the queries, the higher the attack efficiency.


\begin{figure}[t]
    \setlength{\abovecaptionskip}{-0.5cm}
	\begin{center}
		\includegraphics[width=0.8\linewidth]{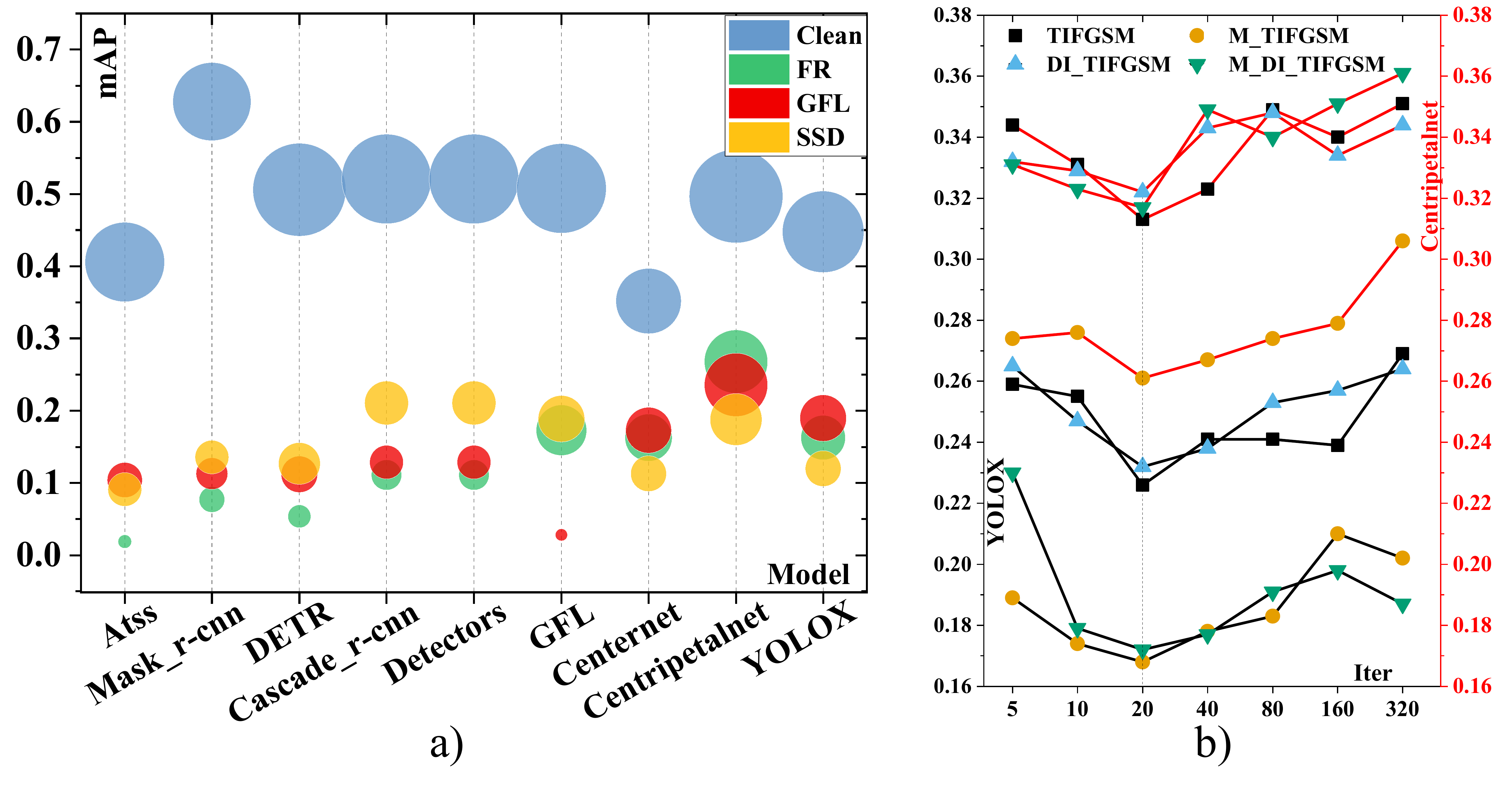}
	\end{center}
	\caption{$\bm{a)}$ We evaluate the performance of transfer attacks generated by three backbone networks on nine models, with the vertical axis representing mAP and the circle radius representing recall. $\bm{b)}$ We verify the effect of different attack algorithms and iterations on the YOLOX and Centripetalnet.}
	\label{fig5}
\end{figure}

\noindent\textbf{Victim Models.} Section 3.2 divides the investigated object detection models into three categories. We selected two black-box models from the modified ResNet series and the self-designed backbone, GFL, Detectors, YOLOX, and Centernet. Among the object models based on the ResNet series backbone, we choose Atss, Casecade R-CNN, Free anchor and Fcos as the black-box attack model. To verify the generation of transferable perturbations, we use Faster R-CNN, GFL and SSD as the white-box attack model and add the transformer-based object detector DETR as the black-box attack model.

\noindent\textbf{Experimental Parameters.} The $w_1$ and $w_2$ are 0.5 in Eq.~\eqref{eq3}. For the selection of random subsets, we use a random search strategy similar to Square attack, sample patches with a size of 0.05 times the original image size in the $weight*height*channel$ subspace as the initial random subset, and initial perturbation of the patch. The sampling size is reduced by half when the queries are [20, 100, 400, 1000, 2000]. In the divide-and-conquer phase, we divide the random subset into four parts and the $i=2$. The population size is set to 2, and the adversarial perturbations respectively generated by Faster R-CNN and SSD iterations 20 times. We set the crossover and mutation rates to 0.8 and 0.3. The norm $n$ is infinity and the budget is 0.05.

\subsection{Transferable Perturbation Generation}
Firstly, we verify the transferability of generative adversarial perturbations on different detectors. We chose three detectors for the white-box attack: Faster R-CNN (FR) based on ResNet50-FPN, GFL based on modified ResNet series, and SSD based on VGG16. In Fig.~\ref{fig5} a), we respectively show the effect of the transfer attack on nine models. The circle's radius represents the mean recall, and the height represents mAP. We have two observations: Firstly, perturbations generated on detectors of the same type perform well. Secondly, adversarial perturbations generated by detectors based on the ResNet can attack most detectors. 

In Fig. 5b), we show the attack effect on Centripalnet(the red axis) and YOLOX(the black axis) of the adversarial perturbations generated by attacking the Faster R-CNN model with different transfer attack methods and different iterations. The M-TIFGSM has the best attack effect. In terms of iterations, the transfer attack has the best effect when about 20 times. As the number of iterations increases, the attack algorithm will gradually overfit. Therefore, we choose M-TIFGSM and iterate 20 times to generate the initial perturbation.

\begin{table}[t]
    \setlength{\belowcaptionskip}{-0.2cm}
	\caption{An ablation study for GARSDC.}
	\label{tab1}
	\footnotesize
	\begin{center}
		\setlength{\tabcolsep}{1pt}{
		\resizebox{9cm}{!}{
			\begin{tabular*}{\textwidth}{ @{\extracolsep{\fill}} lcccccccccccccc}
				\midrule
				\midrule
				\multirow{2}[4]{*}{Method} &\multicolumn{5}{c}{GFL~\cite{li2020generalized}}  & \multicolumn{5}{c}{YOLOX~\cite{ge2021yolox}} \\
				\cmidrule(lr){2-6} \cmidrule(lr){7-11} 
				& $mAP$ & $mAP_{S}$  &  $mAP_{M}$  &  $mAP_{L}$  & $AQ$  & $mAP$ &  $mAP_{S}$  &  $mAP_{M}$  &  $mAP_{L}$  & $AQ$  \\
				\midrule
				\midrule
				Clean & 0.59 & 0.36 & 0.62 & 0.79 & $N/A$ & 0.52 & 0.27 & 0.55 & 0.76 & $N/A$\\
				PRFA & 0.31&	0.17&	0.31&	0.45&  3571&    0.31&	0.15&	0.34&	0.51&	3220 \\
				$\text{PRFA}_{TP+FP}$ & 0.27&	0.16&	0.27&	0.45&  3604&	0.28&   0.11&	0.32&	0.48&	3109\\
				$\text{PRFA}_{TA}$ & 0.21&	0.07&	0.20&   0.33&  3359&	0.31&	0.14&	0.33&	0.50&	3175\\
				$\text{GA}_{TA}$ & 0.25 &	0.07&	0.24&	0.44&  3401&	0.42&	0.17&	0.46&	0.63&	3600\\
				$\text{GARS}_{TA}$ & 0.20&	0.09&	0.20&   0.32&  3133&	0.29&	0.12&	0.32&	0.48&	3201 \\
				$\text{GARSDC}_{TA}$ & 0.18&	0.08&	0.21&	0.32&  3037&    0.31&   0.14&	0.34&	0.50&	3170 \\
				$\text{GARSDC}_{MixTA}$ & $\bm{0.16}$ &	$\bm{0.05}$&	$\bm{0.16}$&	$\bm{0.28}$&  $\bm{1838}$&	$\bm{0.23}$&$	\bm{0.10}$&	$\bm{0.28}$&	$\bm{0.42}$&	$\bm{2691}$\\
				\midrule
				\midrule

\end{tabular*}}}
\end{center}
\end{table}

\subsection{Ablation Study}
To verify the effectiveness of each component of the proposed algorithm, we perform an ablation study on GFL and YOLOX models. $\text{PRFA}_{TP+FP}$ represents replacing the optimization objective of PRFA with `TP+FP'. The subscript ${TA}$ indicates that using the skip-based perturbation as the initial perturbation. GA stands for genetic algorithm for the entire image. GARS stands for Genetic Algorithm with random subset selection. GARSDC stands for Genetic Algorithm based on random subset selection and divide-and-conquer. ${MixTA}$ represents using the skip-based perturbation and chain-based perturbation as the mixed-init populations.

In Tab.~\ref{tab1}, replacing the objective attack function improves the attack effect by 3 points. Replacing the initialization method of PRFA, the improvement of the attack effect is most apparent, which means that our proposed gradient-prior perturbation is better than the previous. The effect of the GA algorithm is not good because the entire image dimension space is too ample for the genetic algorithm, and it is not easy to optimize. After adding random subset selection and divide-and-conquer, the attack performance of the algorithm has been significantly improved (mAP decreased by 7 points in total). After adding the mixed perturbations mechanism, the difference between populations is more significant than that generated by a single model. Consequently, the queries for genetic algorithms are significantly reduced.

\begin{table}[t]
	\setlength{\belowcaptionskip}{-0.2cm}
	\caption{Untargeted attacks against detectors based on different backbones.}
	\label{tab2}
	\footnotesize
	\begin{center}
		\setlength{\tabcolsep}{1pt}{
			\resizebox{9cm}{!}{
			\begin{tabular*}{\textwidth}{ @{\extracolsep{\fill}} lcccccccccccccc}
				\midrule
				\midrule
				\multirow{2}[4]{*}{Method} &\multicolumn{5}{c}{Atss~\cite{zhang2020bridging}}  & \multicolumn{5}{c}{Fcos~\cite{tian2019fcos}} \\
				\cmidrule(lr){2-6} \cmidrule(lr){7-11} 
				& $mAP$ & $mAP_{S}$  &  $mAP_{M}$  &  $mAP_{L}$  & $AQ$  & $mAP$ &  $mAP_{S}$  &  $mAP_{M}$  &  $mAP_{L}$  & $AQ$  \\
				\midrule
				\midrule
				Clean & 0.54&	0.32&	0.58&	0.74&   $N/A$& 	    0.54&	0.33&	0.56&	0.74&	$N/A$\\
				SH & 0.40&	0.20&	0.40&	0.59&  3852&    0.27&	0.09&	0.37&	0.64&	3633 \\
				SQ & 0.23&	0.13&	0.28&	0.31&  3505&    0.21&	0.14&	0.20&	0.37&	3578 \\
				PRFA & 0.20&	0.12&	0.25&	0.30&  3500&    0.23&	0.15&	0.29&	0.41&	3395\\
				GARSDC& $\bm{0.04}$&	$\bm{0.02}$&	$\bm{0.05}$&	$\bm{0.11}$&  $\bm{1837}$&    $\bm{0.15}$&	$\bm{0.09}$&	$\bm{0.17}$&	$\bm{0.28}$&	$\bm{3106}$\\
				\midrule
				\multirow{2}[4]{*}{Method} &\multicolumn{5}{c}{GFL~\cite{li2020generalized}}  & \multicolumn{5}{c}{Centernet~\cite{duan2019centernet}} \\
				\cmidrule(lr){2-6} \cmidrule(lr){7-11} 
				& $mAP$ & $mAP_{S}$  &  $mAP_{M}$  &  $mAP_{L}$  & $AQ$  & $mAP$ &  $mAP_{S}$  &  $mAP_{M}$  &  $mAP_{L}$  & $AQ$  \\
				\midrule
				\midrule
				Clean & 0.59&	0.36&	0.62&	0.79&   $N/A$& 0.44&	0.14&	0.45&	0.71&   $N/A$\\
				SH & 0.43&	0.22&	0.42&	0.59&   3904& 	0.35&	0.08&	0.33&	0.56&  3882 \\
				SQ & 0.33&	0.17&	0.31&	0.50&   3751& 0.25&	0.06&	0.27&	0.44&  3591\\
				PRFA & 0.31&	0.17&	0.31&	0.45&   3570&0.25&	0.07&	0.23&	0.46 & 3697\\
				GARSDC& $\bm{0.16}$&$	\bm{0.05}$&	$\bm{0.16}$&	$\bm{0.28}$&   $\bm{1838}$&   $\bm{0.12}$&	$\bm{0.03}$&	$\bm{0.13}$&$	\bm{0.23}$&  $\bm{2817}$\\
				\midrule
				\multirow{2}[4]{*}{Method} &\multicolumn{5}{c}{YOLOX~\cite{ge2021yolox}}  & \multicolumn{5}{c}{Detectors~\cite{qiao2021detectors}} \\
				\cmidrule(lr){2-6} \cmidrule(lr){7-11} 
				& $mAP$ & $mAP_{S}$  &  $mAP_{M}$  &  $mAP_{L}$  & $AQ$  & $mAP$ &  $mAP_{S}$  &  $mAP_{M}$  &  $mAP_{L}$  & $AQ$  \\
				\midrule
				\midrule
				Clean &0.52&	0.27&	0.56&	0.76&    $N/A$& 	    0.61&	0.39&	0.66&	0.82&    $N/A$\\
				SH & 0.37&	0.15&	0.43&	0.66&    3651& 	0.51&	0.27&	0.48&	0.72&   4000 \\
				SQ & 0.32&	0.17&	0.37&	0.44&    3502&    0.45&	0.23&	0.45&	0.62&	3957\\
				PRFA & 0.31&	0.15&	0.34&	0.51&    3220&    0.41&	0.24&	0.43&	0.58&	3925\\
				GARSDC& $\bm{0.23}$&	$\bm{0.10}$&	$\bm{0.28}$&	$\bm{0.42}$&  $\bm{2691}$&	$\bm{0.28}$&	$\bm{0.09}$&	$\bm{0.27}$&	$\bm{0.49}$&	$\bm{2938}$\\
				\midrule
				\midrule

\end{tabular*}}}
\end{center}
\end{table}

\subsection{Attacks against Detector based on Different Backbones}
In this section, we compare the attack performance of GARSDC and state-of-the-art black-box algorithms on multiple object detectors. In Tab.~\ref{tab2}, we respectively select two object detectors based on three different backbones, which are ATSS based on ResNet101 structure, Fcos based on ResNeXt101 structure, GFL based on ResNeXt101 with deformable convolution, YOLOX based on yolo-s, Centernet based on ResNet18, and Detectors based on RFP and switchable atrous convolution. It is not difficult to see from the experiments that our method reduces by an average 12.0 in the mAP and 980 queries compared with the state-of-the-art algorithm PRFA. The improvement of Atss is the largest, and the attack mAP is 0.04, which may be that our generated skip-based initial perturbation works best to transfer attack against Atss. In addition, our attack effect on Atss, GFL, and Centernet has been improved by more than a half compared with PRFA.

Comparing the size of attack targets, we find that the improvement of our algorithm is mainly focused on small and medium-sized objects. The size of these targets is usually under $64*64$, which is in line with our expectations because the divide-and-conquer method decomposes the random search into smaller search areas, so the attack ability on small and medium objects will be improved. At the same time, the attack of large objects is still difficulty. Comparing the six detectors, Detectors has the most challenging attack (the mAP after the attack is still 0.28), which we think may be related to its structure(switchable atrous convolution), which may inspire us to design a robust architecture for object detection.

\subsection{Visual Analysis}
\begin{figure}[t]
    \setlength{\abovecaptionskip}{-0.5cm}
	\begin{center}
		\includegraphics[width=0.8\linewidth]{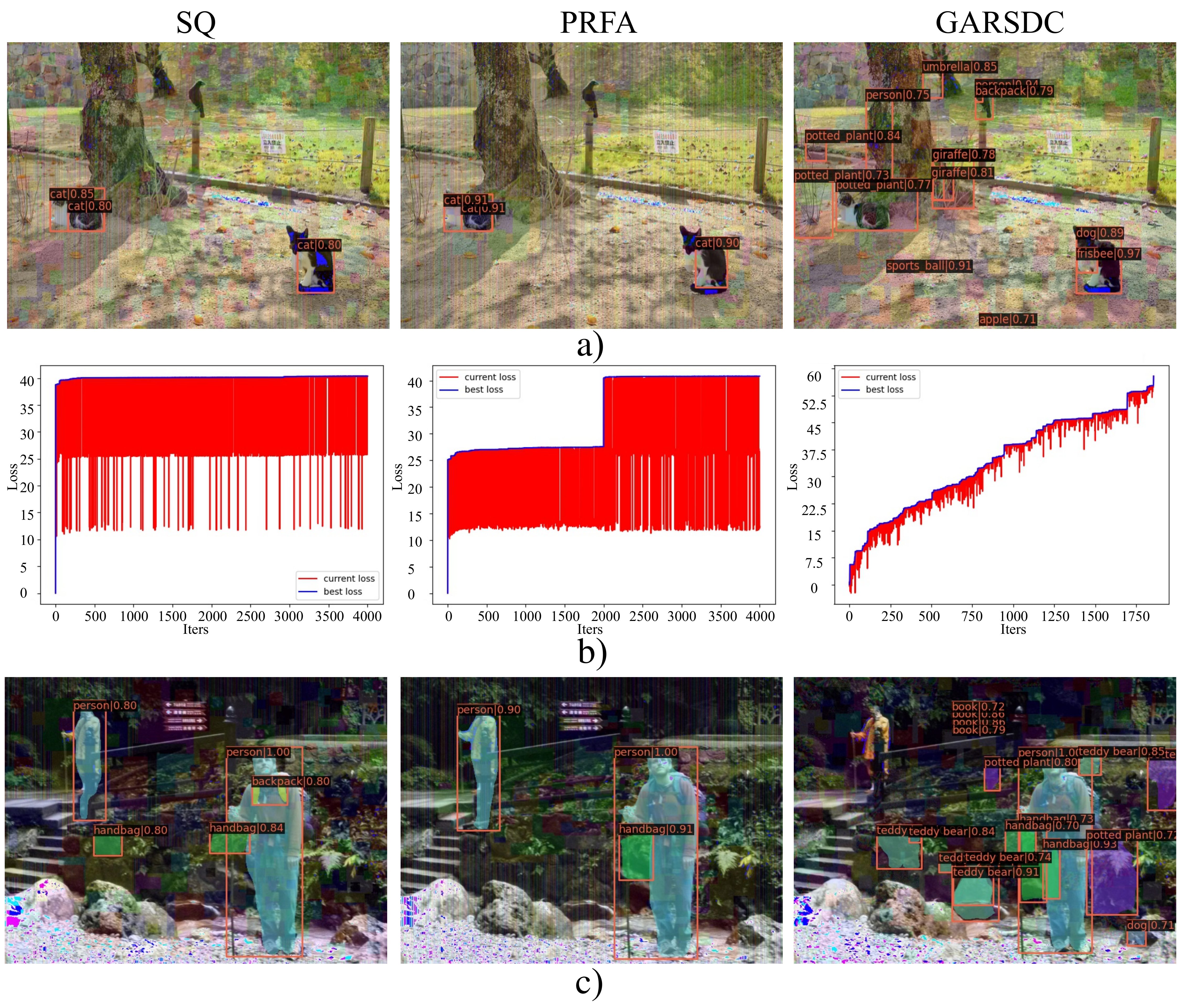}
	\end{center}
	\caption{The SQ~\cite{andriushchenko2020square}, PRFA~\cite{liang2022parallel}, and GARSDC respectively represent the two state-of-art attack methods and our proposed method. We show detection results after attacks in \textbf{a)}, the optimization process of three methods in \textbf{b)} and segmentation results after attacks in \textbf{c)}.}
	\label{fig6}
\end{figure}

This section visualizes the attack results of the square attack, PRFA, and GARSDC. We show the detection results of the three attacks in Fig.~\ref{fig6} a) and the optimization process in Fig.~\ref{fig6} b). During the attack process, we find that GARSDC optimization generates many negative samples and can jump out of local optima during the perturbation iteration process. Both Square attack and PRFA are more likely to fall into local optimal solutions. In Fig.~\ref{fig6} c), we show the detection and segmentation results produced by the three attack methods. We generate multiple small objects and attack pixel classes in a clustered state, which means that the adversarial perturbations we generate can attack the detection and segmentation models.

\section{Conclusion}
In this paper, we model the adversarial attack against object detection as a large-scale multi-objective optimization problem. Unlike the traditional attack method that reduces true positive objects, we minimize the true positive rate and maximize the false positive rate in the attack process to jointly increase the mAP and queries. We propose an efficient genetic algorithm based on random subset selection and divide-and-conquer, optimizing the Pareto-optimal solutions and conquering the challenge of the large-scale decision variables. We generate skip-based and chain-based perturbation by investigating and analyzing more than 40 detection model structures to tackle the problem that the genetic algorithm is sensitive to the population. This gradient-prior population initialization can improve the optimization efficiency of GARSDC. Many attack experiments based on different backbone detectors demonstrate the effectiveness and efficiency of GARSDC. Compared with the state-of-art PRFA algorithm, GARSDC decreases by an average 12.0 in the mAP and queries by nearly 1000 times.

\noindent\textbf{Acknowledgments.} Supported by the National Key R\&D Program of China under Grant 2020YFB1406704, National Natural Science Foundation of China (No. 62025604), Open Project Program of State Key Laboratory of Virtual Reality Technology and Systems, Beihang University (No.VRLAB2021C06). Baoyuan Wu is supported by the Natural Science Foundation of China under grant No.62076213, Shenzhen Science and Technology Program under grants No.RCYX20210609103057050 and No.ZDSYS20211021111415025, and Sponsored by CCF-Tencent Open Fund.
%
%
\bibliographystyle{splncs04}
\bibliography{egbib}
\end{document}


\pagestyle{headings}
\mainmatter
\def\ECCVSubNumber{3147}  

\title{Supplementary Material: A Large-scale Multiple-objective Method for Black-box Attack against Object Detection} 

\titlerunning{GARSDC}
\author{Siyuan Liang\inst{1,2} \and
Longkang Li\inst{3} \and
Yanbo Fan\inst{4} \and
Xiaojun Jia\inst{1,2} \and
Jingzhi Li\inst{1,2, \thanks{Coressponding Author}} \and
Baoyuan Wu\inst{3,\star} \and
Xiaochun Cao \inst{5}}
%
%
\authorrunning{Siyuan Liang, et al.}
\institute{State Key Laboratory of Information Security,
Institute of Information Engineering, Chinese Academy of Sciences, Beijing, China  \and  
School of Cyber Security, University
of Chinese Academy of Sciences, Beijing, China \and
School of Data Science, Secure Computing Lab of Big Data, The Chinese University of Hong Kong, Shenzhen, China  \and
Tencent AI Lab, Shenzhen, China \and
School of Cyber Science and Technology, Shenzhen Campus, Sun Yat-sen University, Shenzhen, China \\
\email{\{liangsiyuan, jiaxiaojun, lijingzhi\}@iie.ac.cn};
\email{\{lilongkang, wubaoyuan\}@cuhk.edu.cn};
\email{fanyanbo0124@gmail.com};
\email{caoxiaochun@mail.sysu.edu.cn} 
}  

\begin{comment}

\end{comment}
\maketitle

\section*{Organization of the Supplementary Material}
In Section A, we show the algorithm flow of GARSDC, which attacks against object detection. In Section B, we add the specific proof for the lower bound(Eq. (8)) of the $P$ values of the best $\boldsymbol{b}_{j}$. Section C gives the approximation guarantee of GARSDC(Eq. (9)). 

\section{The Algorithm Flow of GARSDC}
\begin{breakablealgorithm}
\label{alg1}
\caption{Adversarial Examples Generation with GARSDC Attack}
\begin{algorithmic}[1] 
\REQUIRE
victim detector $H_1$, gradient-prior detectors $H_2$,$H_3$, clean image $\boldsymbol{x}\in\mathbb{R}^{w \times h \times c}$, ground-truth boxes $\mathcal{O}$, maximum number of iterations $T_{max}$, the size of
a sample patch $a$, maximum iterations $T_{dc}$ of divide-and-conquer, the flag of divide-and-conquer $flag$, numbers of divide-and-conquer $i$, crossover rate $cr$, mutation rate $mr$.
\ENSURE adversarial image $\hat{\boldsymbol{x}}\in\mathbb{R}^{w \times h \times c}$ with $||\boldsymbol{\delta}||_{\infty}\leq 0.05$, $\hat{\boldsymbol{x}}=\boldsymbol{x}+\boldsymbol{\delta}$
\\
$\boldsymbol{\delta}_1^{best}, \boldsymbol{\delta}_2^{best} \leftarrow$ INIT$(\boldsymbol{x}, H_2, H_3)$(see Alg. 2 for the initial population in Section 3.3)
\STATE $f_{best}\leftarrow \max \{F(H_{1}(\boldsymbol{x}), \boldsymbol{x}+\boldsymbol{\delta}_m)\}, m = 1, 2$ in Eq. (3) with ground-truth boxes $\mathcal{O}$
\STATE $S_{best}=\{\infty\}_{j=1}^{i}$ 
\STATE Best individual's index is $m\leftarrow \arg \max \limits_{\boldsymbol{\delta}_m} \{F(H_{1}(\boldsymbol{x}), \boldsymbol{x}+\boldsymbol{\delta}_m)\}, m=1,2$
\STATE $t \leftarrow 2, flag \leftarrow 0, a^{(2)} \leftarrow 0.05 * min(w, h)$
\WHILE{$t<T_{max}$ \textbf{and} $\hat{\boldsymbol{x}}$ is not adversarial}
\STATE $a^{(t)} \leftarrow $side length of the sample patch (according to some schedule)
\STATE $\boldsymbol{\delta}_1^{new}$, $\boldsymbol{\delta}_2^{new}$, $\boldsymbol{s}=$ RS($w, h, c, \boldsymbol{\delta}_1^{best}, \boldsymbol{\delta}_2^{best}, a^{(t)}$)(see Alg. 3 for Section 3.4)
\STATE Partition $\boldsymbol{s}$ into $i$ sets $\boldsymbol{s}_1$, $\boldsymbol{s}_2$, ..., $\boldsymbol{s}_{i}$ evenly so that each $\boldsymbol{s}_j$ is a square with equal sides
\STATE $f^{new}_{m}, S_{new}^{m} \leftarrow  F(H_{1}(\boldsymbol{x}), \boldsymbol{x}+\boldsymbol{\delta}_m^{new}), S(\boldsymbol{\delta}_m^{new}[\boldsymbol{s}])$
\STATE $t \leftarrow t+1$
\IF {$\exists s_j \in S_{new}, s_j \neq \infty $} $flag=1$
\ENDIF

\IF {$flag == 1$}
\WHILE{$j<i$}
\STATE // Run Alg. 4 with $	T=T_{dc}-1$ on each $\boldsymbol{s}_j$
\STATE $\boldsymbol{\delta}_1[\boldsymbol{u}_j], \boldsymbol{\delta}_2[\boldsymbol{u}_j], f(\boldsymbol{\delta}_{1j}), f(\boldsymbol{\delta}_{2j})=$ DC($\boldsymbol{\delta}_1^{new}, \boldsymbol{\delta}_2^{new}, T_{dc}-1, H_1, \mathcal{O}, mr, cr, \boldsymbol{s}_j, \boldsymbol{x}$)
\ENDWHILE
\STATE Merge the $i$ resulting subsets into a set $U=\bigcup_{j=1}^{i}\boldsymbol{u}_j$
\STATE // Run Alg. 4 with $	T=1$ on each $U$
\STATE $\boldsymbol{\delta}_1[\boldsymbol{u}_{i+1}], \boldsymbol{\delta}_2[\boldsymbol{u}_{i+1}], f(\boldsymbol{\delta}_{1({i+1})}), f(\boldsymbol{\delta}_{2({i+1})})=$ DC($\boldsymbol{\delta}_1^{new}, \boldsymbol{\delta}_2^{new}, 1, H_1, \mathcal{O}, mr, cr, U, \boldsymbol{x}$)
\STATE The best sub-component is $\boldsymbol{\delta}[\boldsymbol{u}_{best}]=\arg \max\limits_{\boldsymbol{\delta}[\boldsymbol{u}_j]} \{f(\boldsymbol{\delta}[\boldsymbol{u}_j])\}, j=1,...,i+1$
\STATE Update $\boldsymbol{\delta}_{1}^{new}, \boldsymbol{\delta}_{2}^{new}$ with the best sub-component $\boldsymbol{\delta}_1[\boldsymbol{u}_{best}], \boldsymbol{\delta}_2[\boldsymbol{u}_{best}]$
\STATE Update individual's fitness $f_{1}^{new}$ and $f_{2}^{new}$
\STATE Update best individual's index $m\leftarrow \arg \max (f_{1}^{new}, f_{2}^{new})$
\STATE $flag \leftarrow 0, t=t+T_{dc}$
\ENDIF
\IF {$f^{new}_{m}>f_{best}$} $\boldsymbol{\delta}_{m}^{best} \leftarrow \boldsymbol{\delta}_{m}^{new}, f_{best} \leftarrow f^{new}_{m}$ 
\ENDIF
\ENDWHILE
\\
return adversarial perturbation $\boldsymbol{\delta}_{m}^{best}$
\end{algorithmic}
\label{al1}
\end{breakablealgorithm}
We show the overall algorithm of GARSDC in Alg.~\ref{al1}. In line 1, we use gradient-prior detectors consisting of Faster r-cnn $H_2$ and SSD $H_3$ to generate skip-based and chain-based perturbations as initial population, which is called mixed initial population in Section 3.3. In line 8, we randomly sample subsets across the full image using a strategy similar to PRFA. However, we do not need a prior-guided dimension reduction, which can circumvent the risk that the prior is terrible. In lines 12-13, we can use the sub-component fitness to help judge whether the sub-components of random search are helpful for optimization and decide search perturbation locally. In line 15, we simultaneously run Alg. 4 on $i$ sets. In the first round (line 17), it evenly distributes the ground set $\boldsymbol{s}$ over $i$ machines, and then each machine runs DC to find a subset $\boldsymbol{s}_j$ in parallel. In the second round, we merge the $i$ resulting subsets are merged on one machine (line 19), and then DS to find another $\boldsymbol{s}_{i+1}$ (line 21). We update $\boldsymbol{\delta}_1^{new}, \boldsymbol{\delta}_2^{new}$ with the best sub-component and record individual fitness $f_1^{new}, f_2^{new}$ at the same time.

\begin{breakablealgorithm}
\caption{Mixed Initial Population Based on Gradient-prior}
\begin{algorithmic}[1] 
\REQUIRE
gradient-prior detectors $H_2$,$H_3$, clean image $\boldsymbol{x}\in\mathbb{R}^{w \times h \times c}$.
\ENSURE adversarial image $\hat{\boldsymbol{x}}\in\mathbb{R}^{w \times h \times c}$ with $||\boldsymbol{\delta}||_{\infty}\leq 0.05$, $\hat{\boldsymbol{x}}=\boldsymbol{x}+\boldsymbol{\delta}$
\STATE $\boldsymbol{\delta}_1^{(0)} \leftarrow 0$, $\boldsymbol{\delta}_2^{(0)} \leftarrow 0$, $t \leftarrow 0$
\WHILE{$t<20$}
\STATE // skip-based perturbation for Eq. (5)
\STATE Input $\boldsymbol{x}+\boldsymbol{\delta}_1^{(t)}$ to detector $H_2$ and obtaion the gradient $ \boldsymbol{W} \ast \nabla_{\boldsymbol{x}}H_2(\boldsymbol{x}+\boldsymbol{\delta}_1^{(t)})$
\STATE 
\STATE Update $\boldsymbol{\delta}_1^{(t+1)}$ by $0.05 \cdot \sign (\boldsymbol{W} \ast \nabla_{\boldsymbol{x}}H_3(\boldsymbol{x}+\boldsymbol{\delta}_1^{(t)}))$
\STATE // chain-based perturbation for Eq. (5)
\STATE Input $\boldsymbol{x}+\boldsymbol{\delta}_2^{(t)}$ to detector $H_3$ and obtaion the gradient $ \boldsymbol{W} \ast \nabla_{\boldsymbol{x}}H_3(\boldsymbol{x}+\boldsymbol{\delta}_2^{(t)})$
\STATE Update $\boldsymbol{\delta}_2^{(t+1)}$ by $0.05 \cdot \sign (\boldsymbol{W} \ast \nabla_{\boldsymbol{x}}H_3(\boldsymbol{x}+\boldsymbol{\delta}_2^{(t)}))$
\ENDWHILE
\STATE return perturbations $\boldsymbol{\delta}_{1}^{(19)}, \boldsymbol{\delta}_{2}^{(19)}$
\end{algorithmic}
\label{alg2}
\end{breakablealgorithm}

We take the TIFGSM attack method as an example, and in Alg.~\ref{alg2}, we show the process that TIFGSM iterates 20 times on the detectors to generate the initialization perturbation. In Alg.~\ref{alg2}, $\boldsymbol{W}$ denotes the kernel matrix in TIFGSM attack. We can return results skip-based $\boldsymbol{\delta}_{1}$ and chain-based perturbations $\boldsymbol{\delta}_{2}$ as the mixed initial population with gradient-prior.

\begin{breakablealgorithm}
\caption{Random Subset Selection}
\begin{algorithmic}[1] 
\REQUIRE
image width $w$, height $h$, channels $c$, skip-based perturbation $\boldsymbol{\delta}^1$, chain-based perturbation $\boldsymbol{\delta}^2$, random subset's size $a$
\STATE $\boldsymbol{s} \leftarrow$ array of zeros of size $w \times h \times c$
\STATE sample uniformly $r \in \{0,...,w-a\}, s \in \{0,...,h-a\}$
\STATE $\boldsymbol{s}_{r+1:r+a, s+1:s+a} = 1$
\WHILE {$i<c$}
\STATE $\rho_1 \leftarrow Uniform(\{-1, 1\}), \rho_2 \leftarrow Uniform(\{-1, 1\})$
\STATE $\boldsymbol{\delta}^{1}_{r+1:r+a, s+1:s+a} \leftarrow (\rho_1 \cdot  \boldsymbol{\delta}^{1})_{r+1:r+a, s+1:s+a}$
\STATE $\boldsymbol{\delta}^{2}_{r+1:r+a, s+1:s+a} \leftarrow (\rho_2 \cdot  \boldsymbol{\delta}^{2})_{r+1:r+a, s+1:s+a}$
\ENDWHILE
\STATE return new perturbations $\boldsymbol{\delta}^1, \boldsymbol{\delta}^2$, and subset $\boldsymbol{s}$
\end{algorithmic}
\label{alg3}
\end{breakablealgorithm}

In Alg.~\ref{alg3}, we show random subset selection. In the first round, we sample subset in the search space via random search (line 2). In the second round, we record the sample subset (line 3) and alter perturbations $\boldsymbol{\delta}^1, \boldsymbol{\delta}^2$ (lines 6-7).

\begin{breakablealgorithm}
\caption{Divide-and-Conquer Algorithm}
\begin{algorithmic}[1] 
\REQUIRE
skip-based perturbation $\boldsymbol{\delta}^{0}_{1}$, chain-based perturbation $\boldsymbol{\delta}^{0}_{2}$, maximum iterations T, victim detector $H_1$, ground-truth boxes $\mathcal{O}$, crossover rate $cr$, mutation rate $mr$, sub-component $\boldsymbol{s}$, clean image $\boldsymbol{x}$
\ENSURE adversarial image $\hat{\boldsymbol{x}}\in\mathbb{R}^{w \times h \times c}$ with $||\boldsymbol{\delta}||_{\infty}\leq 0.05$, $\hat{\boldsymbol{x}}=\boldsymbol{x}+\boldsymbol{\delta}$
\STATE // genetic algorithm
\STATE $f_1^{(0)}, f_2^{(0)}, s_1^{(0)}, s_2^{(0)} \leftarrow F(H_1(\boldsymbol{x}), \boldsymbol{x}+\boldsymbol{\delta}_1), F(H_1(\boldsymbol{x}), \boldsymbol{x}+\boldsymbol{\delta}_2), S(\boldsymbol{\delta}_1[\boldsymbol{s}]), S(\boldsymbol{\delta}_2[\boldsymbol{s}])$ with ground-truth boxes $\mathcal{O}$ in Eq. (3) and Eq. (7).
\STATE $f_1^{bset}, f_2^{best}, \boldsymbol{\delta}_1^{best}[\boldsymbol{s}], \boldsymbol{\delta}_1^{best}[\boldsymbol{s}] \leftarrow f_1^{0}, f_2^{0}, \boldsymbol{\delta}_1^{0}[\boldsymbol{s}], \boldsymbol{\delta}_1^{0}[\boldsymbol{s}]$
\STATE $t \leftarrow 0$
\WHILE{$t<T$}
\IF{$\max \{s_1^{(t)}, s_2^{(t)}\} = \infty$} 
\STATE break
\ENDIF
\STATE $\rm{loser, winner} = \rm{sort\_by\_fitness}(\boldsymbol{\delta}_1^{(t)}[\boldsymbol{s}], \boldsymbol{\delta}_2^{(t)}[\boldsymbol{s}], s^{(t)}_1, s^{(t)}_2)$
\STATE $\boldsymbol{\delta}_1^{(t)}[\boldsymbol{s}], \boldsymbol{\delta}_2^{(t)}[\boldsymbol{s}] = \rm{crossover}(cr, \rm{loser, winner})$
\STATE $\boldsymbol{\delta}_1^{(t)}[\boldsymbol{s}], \boldsymbol{\delta}_2^{(t)}[\boldsymbol{s}] = \rm{mutation}(mr, \rm{loser, winner})$
\STATE $t \leftarrow t+1$
\STATE $f_1^{(t)}, f_2^{(t)}, s_1^{(t)}, s_2^{(t)} \leftarrow F(H_1(\boldsymbol{x}), \boldsymbol{x}+\boldsymbol{\delta}_1^{(t)}), F(H_1(\boldsymbol{x}), \boldsymbol{x}+\boldsymbol{\delta}_2^{(t)}), S(\boldsymbol{\delta}_1^{(t)}[\boldsymbol{s}]), S(\boldsymbol{\delta}_2^{(t)}[\boldsymbol{s}])$ with ground-truth boxes $\mathcal{O}$ in Eq. (3) and Eq. (7).
\IF{$f_1^{(t)}>f_1^{(t-1)}$}
\STATE $f_1^{best} \leftarrow f_1^{t}, \boldsymbol{\delta}_1^{best}[\boldsymbol{s}] \leftarrow \boldsymbol{\delta}_1^{t}[\boldsymbol{s}]$
\ENDIF
\IF{$f_2^{(t)}>f_2^{(t-1)}$}
\STATE $f_2^{best} \leftarrow f_2^{t}, \boldsymbol{\delta}_2^{best}[\boldsymbol{s}] \leftarrow \boldsymbol{\delta}_2^{t}[\boldsymbol{s}]$
\ENDIF
\ENDWHILE
\STATE return best sub-components $\boldsymbol{\delta}_1^{best}[\boldsymbol{s}],\boldsymbol{\delta}_2^{best}[\boldsymbol{s}]$ and individual's fitness $f_1^{best}, f_2^{best}$
\end{algorithmic}
\label{alg4}
\end{breakablealgorithm}
In Alg.~\ref{alg4}, we show the genetic algorithm based on the divide-and-conquer algorithm. In line 6, it denotes there are no true positive or false positive objects in $\boldsymbol{s}$. Thus, the sub-component do not need to be improved. We sort the fitness of sub-components in line 9 and alter sub-components by crossover and mutation in lines 10-11. Since object detection is holistic, we return the best fitness of individuals and corresponding sub-components over the entire process.
\section{Proof A Lower Bound on The $P$ Value}
In Section 3.2, we set the $p$-th individual fitness $P(\boldsymbol{x}+\boldsymbol{\delta}_p)=F(\boldsymbol{x}+\boldsymbol{\delta}_p)$. Thus, proof a lower bound on the $P$ value is same as on the $F$ value. We firstly give two notions of `approximate submodularity', which measure to the extent of a set function $F$ has the submodular property. The $\gamma$- and $\alpha$-submodularity as follows: 

\textbf{Definition 1} $\gamma$-Submodularity Ratio~\cite{das2011submodular}. The submodularity ratio of a set function $F:2^{\boldsymbol{s}} \rightarrow \mathbb{R}$ with respect to a set $\boldsymbol{u} \subseteq \boldsymbol{s}$ and a paremeter $l \geq 1$ is
\begin{equation}
	\gamma_{\boldsymbol{u}, l}(F) = \min \limits_{\boldsymbol{l} \subseteq \boldsymbol{u},\boldsymbol{m}:|\boldsymbol{m}|\leq l, \boldsymbol{m} \cap \boldsymbol{l} = \emptyset}\frac{\sum\nolimits_{v \in \boldsymbol{m}}(F(\boldsymbol{l} \cup v)-F(\boldsymbol{l}))}{F(\boldsymbol{l} \cup \boldsymbol{m})-F(\boldsymbol{l})} \tag{10}
\label{eq10}
\end{equation}

\textbf{Definition 2} $\alpha$-Submodularity Ratio~\cite{qian2016parallel}. The submodularity ratio of a set function $F:2^{\boldsymbol{s}} \rightarrow \mathbb{R}$ is
\begin{equation}
	\alpha_{F} = \min \limits_{\boldsymbol{u} \subseteq \boldsymbol{m} \subseteq \boldsymbol{s}, v \notin \boldsymbol{m}}\frac{F(\boldsymbol{s} \cup v)-F(\boldsymbol{s})}{F(\boldsymbol{m} \cup v)-F(\boldsymbol{m})} \tag{11}
\end{equation}
The subset selection problem as follows:

\textbf{Definition 3} Subset Selection. Given all items $\boldsymbol{s}=\{s_1,...,s_n\}$, an objective function $F$ and a budget $z$. we aim to find a subset of most $z$ items maximizing $F$, i.e.,
\begin{equation}
	\arg \max \nolimits_{\boldsymbol{u} \subseteq \boldsymbol{s}}F(\boldsymbol{u}) , s.t. |\boldsymbol{u}| \leq z \tag{12}
\end{equation}
where $|\cdot|$ denotes the size of a set. 

Now, we will proof the Eq. (8). We first prove that $\max \{P(\boldsymbol{\delta}[\boldsymbol{b}_j])|1 \leq j \leq i \} \geq \frac{\alpha}{i}OPT$. The $\boldsymbol{b}$ denotes an optimal subset of $\boldsymbol{s}$, i.e., $P(\boldsymbol{\delta}[\boldsymbol{b}])=OPT$. For $1 \leq j \leq i$, Let $\boldsymbol{a}_{j}=\boldsymbol{b} \cap \boldsymbol{s}_j$. Thus, $\cup_{j=1}^i \boldsymbol{a}_{j}=\boldsymbol{b}$ and for any $j \neq m$, $\boldsymbol{a}_{j} \cap \boldsymbol{a}_{m} = \emptyset$. Then, we have
\begin{equation}
	P(\boldsymbol{\delta}[\boldsymbol{b}])= P(\boldsymbol{\delta}[\cup_{j=1}^i \boldsymbol{a}_{j}])= \sum_{j=1}^{i}P(\boldsymbol{\delta}[\cup_{m=1}^j \boldsymbol{a}_{m}])-P(\boldsymbol{\delta}[\cup_{m=1}^{j-1} \boldsymbol{a}_{m}]) \tag{13}
\end{equation}
The set $\{s_1^{j},...,s_{|\boldsymbol{a}_{j}|}^{j}\}$ denotes the items in $\boldsymbol{a}_{j}$. Then, for any $1 \leq j \leq i$, it holds that
\begin{equation}
\begin{aligned}
	P(\boldsymbol{\delta}[\cup_{m=1}^j \boldsymbol{a}_{m}])-P(\boldsymbol{\delta}[\cup_{m=1}^{j-1}\boldsymbol{a}_{m}]) \qquad \qquad \qquad \qquad \qquad \qquad \qquad \qquad \qquad & \\= \sum_{l=1}^{|\boldsymbol{a}_{j}|}P(\boldsymbol{\delta}[\cup_{m=1}^{j-1}\boldsymbol{a}_{m} \cup \{s_1^{j},...,s_l^{j}\}]) -  P(\boldsymbol{\delta}[\cup_{m=1}^{j-1}\boldsymbol{a}_{m} \cup \{s_1^{j},...,s_{l-1}^{j}\}]) \\ \leq \frac{1}{\alpha}\sum_{l=1}^{|\boldsymbol{a}_{j}|}P(\boldsymbol{\delta}[\{s_1^{j},...,s_l^{j}\}])- P(\boldsymbol{\delta}[\{s_1^{j},...,s_{l-1}^{j}\}]) = \frac{P(\boldsymbol{\delta}[\boldsymbol{a}_{j}])}{\alpha}, \qquad \qquad 
\end{aligned}
\tag{14}
\end{equation}
where the inequality is by the definition of $\alpha$-submodularity ratio since $\{s_1^{j},...,s_{l-1}^{j}\} \subseteq \cup_{m=1}^{j-1}\boldsymbol{a}_{m} \cup \{s_1^{j},...,s_{l-1}^{j}\}$. Note that for any $1 \leq j \leq i, P(\boldsymbol{\delta}[\boldsymbol{b}_{j}]) \geq P(\boldsymbol{\delta}[\boldsymbol{a}_{j}])$, since $\boldsymbol{a}_{j} \subseteq \boldsymbol{s}_{j}$ and $|\boldsymbol{a}_{j}| \leq |\boldsymbol{b}| \leq z$. Thus we get
\begin{equation}
	OPT=P(\boldsymbol{a}[\boldsymbol{b}]) \leq \frac{1}{\alpha} \sum_{j=1}^{i}P(\boldsymbol{a}_{j}) \leq \frac{1}{\alpha} \sum_{j=1}^{i}P(\boldsymbol{b}_{j}) \tag{15}
\end{equation}
which leads to $\max \{P(\boldsymbol{\delta}[\boldsymbol{b}_j])|1 \leq j \leq i \} \geq \frac{\alpha}{i}OPT$. 

We then prove that $\max \{P(\boldsymbol{\delta}[\boldsymbol{b}_j])|1 \leq j \leq i \} \geq \frac{\gamma_{\emptyset, z}}{z}OPT$. By the definition of $\gamma$-submodularity ratio, $P(\boldsymbol{\delta}[\boldsymbol{b}]) \leq \sum_{s \in \boldsymbol{b}} F(\boldsymbol{\delta}[s])/\gamma_{\emptyset, z}$. Let $s^{*} \in \arg \max \nolimits_{s \in \boldsymbol{b}} P(\boldsymbol{\delta}[s])$, and $\{\boldsymbol{s}_1,...,\boldsymbol{s}_i\}$ is a partition set of $\boldsymbol{s}$, $s^{*} $ must belong to one of the $j$-th sets. Thus $\max \{P(\boldsymbol{\delta}[\boldsymbol{b}_j])|1 \leq j \leq i \} \geq \frac{\gamma_{\emptyset, z}}{z}OPT$. We put $z$ into the above equation and Eq. (15) and proof the Eq. (8)

\section{Proof Approximation Guarantee}
For any $\boldsymbol{u} \subseteq \boldsymbol{s}_j(1 \leq j \leq i)$, there exits one item $s \in \boldsymbol{s}_j / \boldsymbol{u}$ such that~\cite{das2011submodular}
\begin{equation}
P(\boldsymbol{\delta}[\boldsymbol{u} \cup s])-P(\boldsymbol{\delta}[s]) \geq (\gamma_{\boldsymbol{u},z} / z)*(P(\boldsymbol{\delta}[\boldsymbol{b}_j])-P(\boldsymbol{\delta}[s]))
\tag{16}
\end{equation} 

In above section, we analyze the maximum number of iterations on each machine until $P(\boldsymbol{\delta}[\boldsymbol{u}]) \geq (1-e^{-\gamma_{min}}) \cdot P(\boldsymbol{\delta}[\boldsymbol{o}_{j}])$. For the machine running DS on $\boldsymbol{s}_j$, let $J_{max}^{j}$ denote the maximum value of $m \in \{1,...,z\}$ such that in the archive $\rm{\Omega}$, there exists a solution $\boldsymbol{u}$  with $\boldsymbol{u} \leq j$ and $P(\boldsymbol{\delta}[\boldsymbol{u}]) \geq (1-(1-\frac{\gamma_{min}}{z})) \cdot P(\boldsymbol{\delta}[\boldsymbol{o}_{j}])$. That is,
\begin{equation}
	J_{max}^{j} = max\{ m \in \{1,...,k\} | \exists \boldsymbol{u} \in R, |\boldsymbol{u}| < k \wedge P(\boldsymbol{\delta}[\boldsymbol{u}]) \geq (1-(1-\frac{\gamma_{min}}{z})) \cdot P(\boldsymbol{\delta}[\boldsymbol{o}_{j}]) \}
\tag{17}
\end{equation}
Then, we only need to analyze the maximum iterations until $\min \{ J_{max}^{j}|1 \leq j \leq i \}=z$. For GA running on $\boldsymbol{s}_j$, let $\boldsymbol{z}_j$ be a corresponding solution with the value $J_{max}^{j}$. Because $\boldsymbol{z}_j$ is weakly dominated by the newly generated solution, the $J_{max}^{j}$ can not decrease. In Eq. (16), we know that for any $1 \leq j \leq i$, change specific bit of $\boldsymbol{z}_j$ can generate a new solution $\boldsymbol{s}_j^{new}$ and satisfy that $P(\boldsymbol{\delta}[\boldsymbol{s}_j^{new}])-P(\boldsymbol{\delta}[\boldsymbol{z}_j]) \geq \frac{\gamma_{\boldsymbol{z}_j, z}}{z}(P(\boldsymbol{\delta}[\boldsymbol{b}_j])-P(\boldsymbol{\delta}[\boldsymbol{z}_j]))$. Then, if $J_{max}^{j}<z$, we have
\begin{equation}
	P(\boldsymbol{\delta}[\boldsymbol{s}_j^{new}]) \geq (1-(1-\gamma_{min}/z)^{J_{max}^{j}+1}) \cdot P(\boldsymbol{\delta}[\boldsymbol{b}_j])
\tag{17} 
\end{equation}
where $\gamma_{\boldsymbol{z}_j, z} > \gamma_{min}$. Since $|\boldsymbol{s}_j^{new}|=|\boldsymbol{z}_j|+1 \leq J_{max}^{j}+1$, $\boldsymbol{s}_j^{new}$ will be included into $\rm{\Omega}$. If $\boldsymbol{s}_j^{new}$ may be dominated by one solution in $\rm{\Omega}$, which are contradictory with the $J_{max}^{j}$. Due to that $\boldsymbol{s}_j^{new}$, $J_{max}^{j}$ increases by at least 1 and the $\rm{\Omega}_{max}$ denotes the largest size of $\rm{\Omega}$ in the Alg.~\ref{alg4}, the $J_{max}^{j}$ can increase by at least 1 in one iteration with probability $\frac{1}{e n_{j} \rm{\Omega}_{max} }$. The $\frac{1}{\rm{\Omega}_{max}}$ is a lower bound on the crossover probability of selecting $\boldsymbol{z}_j$ and $n_{j}$ is the mutation rate. Because that solution limitation, we have $\rm{\Omega}_{max} \leq 2z$. We the get that after one iteration in the first around DC, $l$ can decrease by at least 1 with probability as least 
\begin{equation}
	1- \prod_{j:J_{max}^{j}=m}(1-\frac{1}{2ez n_{m}}) \geq 1-(1-\frac{1}{2ez n_{max}})^{l}
\tag{18}
\end{equation}
since it is sufficient that at least one of those $J_{max}^{j}=m$ increases. Thus, the expected number of iterations until $m$ increases
\begin{equation}
	\sum_{l=1}^{i} \frac{1}{1-(1-\frac{1}{2ez n_{max}})^{l}}\leq i+(2ezn_{max}-1)H_i
\tag{19}
\end{equation}
where $H_i$ is the $j$-th harmonic number. Then the complexity is $O(z^2n_{max}(1+ \log i))$. Since $\min \{J^{j}_{max}|1 \leq j \leq i\}=z$ implies that $P(\boldsymbol{\delta}[\boldsymbol{u}]) \geq (1-e^{-\gamma_{min}}) \cdot P(\boldsymbol{\delta}[\boldsymbol{o}_{j}])$ for any $1 \leq j \leq i$, the $P$ value of the final output subset satisfies that $\max \{P(\boldsymbol{\delta}[\boldsymbol{u}_j])| 1 \leq j \leq i \} \geq (1-e^{-\gamma_{min}}) * \max\{ P(\boldsymbol{\delta}[\boldsymbol{b}_j])| 1 \leq j \leq i \}$. By Lemma 1, Eq. (9) holds.

%
%
\bibliographystyle{splncs04}
\bibliography{egbib}